\documentclass{article}
\usepackage{amsmath,graphicx}
\usepackage{INTERSPEECH2022}

\usepackage{multirow}
\usepackage{color}
\usepackage[dvipsnames]{xcolor}
\usepackage[normalem]{ulem}

\newcommand{\yv}[1]{\textcolor{black}{#1}} %RedOrange

\title{Isochrony-Aware Neural Machine Translation for Automatic Dubbing}

\name{Derek Tam*\thanks{*This work was done during an internship at Amazon.}, Surafel M. Lakew, Yogesh Virkar, Prashant Mathur, Marcello Federico}
\address{\yv{AWS AI Labs}}
\email{\small \tt \{surafelm|yvvirkar|pramathu|marcfede\}@amazon.com}
\begin{document}
\ninept
\maketitle
\begin{abstract}
We introduce the task of isochrony-aware machine translation which aims at generating translations suitable for dubbing. 
Dubbing of a spoken sentence requires transferring the content as well as the speech-pause structure of the source into the target language to achieve audiovisual coherence. Practically, this implies correctly projecting pauses from the source to the target and ensuring that target speech segments have roughly the same duration of the corresponding source speech segments. In this work, we propose implicit and explicit modeling approaches to integrate isochrony information into neural machine translation. Experiments on English-German/French language pairs with automatic metrics show that the simplest of the considered approaches works best. Results are confirmed by human evaluations of translations and dubbed videos.
\end{abstract}

\yv{\noindent\textbf{Index Terms}: Machine Translation, Isochrony, Prosody, Verbosity, Automatic Dubbing}

\section{Introduction}
\label{sec:intro}
% motivate PA MT 
Recent advancements in machine translation (MT), largely due to the success of transformer models, have improved the quality of MT significantly~\cite{vaswani2017attention}. However, when MT is applied to specific use cases, like subtitle translation or automatic dubbing~\cite{oktem2019,federico_speech--speech_2020,federico_evaluating_2020}, translation quality is not the only dimension by which a model's performance is evaluated. In subtitles, translation of a source sentence should fit in a fixed block size \cite{karakanta_is_2020}. 
Automatic Dubbing requires \textit{isochrony} i.e. when the character is on-screen the translation (of source \yv{utterance} line) should match the timing of original speech and speech-pause temporal arrangement in the original audio~\cite{virkar2021pa}. This means, pauses in the source \yv{utterance} should be projected into the target \yv{translation} in relatively similar positions~\cite{federico2020eval}. 
In this paper, our focus is on {translation and projection of pause markers} in the correct position to \textbf{enable isochrony} in dubbing.

\yv{Currently in an automatic dubbing pipeline~\cite{oktem2019,federico_speech--speech_2020,federico_evaluating_2020}, a source utterance is first \textit{translated} and then a prosodic alignment (PA) model~\cite{virkar2021pa} segments the translated text into phrases and pauses following the phrase-pause arrangement of the source utterance~\cite{federico_evaluating_2020, virkar2021pa}}.\footnote{In this paper, following~\cite{virkar2021pa} we define pause as 300ms of silence between two consecutive spoken words. \yv{We define a phrase (or interchangeably a segment) as the text between two pauses.}} 
In these two steps, two distinct models are deployed, one for translation and one for segmentation, which is clearly a sub-optimal solution. Our hypothesis is that better and more suitable translations could be generated by taking into account the phrase-pause structure to be targeted.

In this paper, we propose to combine the two steps into a single MT model that directly generates translations including pause markers. 
\yv{We, therefore, introduce a new task of Isochrony-Aware MT (IAMT) where MT system should jointly transfer both meaning and the phrase-pause structure 
from source to target language.}

The task of IAMT is challenging in different ways: 1) MT needs to learn two distinct modeling problems: MT and PA; 2) while learning to project pauses, MT should not deteriorate translation quality; 3) MT should temporally map a source segment (text between two pauses) into a target segment of similar duration. In particular, the third challenge requires MT
to control the verbosity of translation at the segment level rather than just at the sentence level. %~\cite{lakew2021verbosity}.

As part of recent efforts to achieve a better synchrony between source and target speech, most of the works have been focused on controlling the length of translated text i.e. its verbosity. \cite{Lakew19} introduced a prefix verbosity control token to control for length and later \cite{lakew2021verbosity} extended the same by generating multiple length controlled hypotheses and rescoring them according to a synchrony score~\cite{saboo-baumann-2019-integration}. \cite{Lakew19,niehues_machine_2020} controlled the verbosity by utilizing positional encoding in the transformer architecture~\cite{vaswani2017attention} while \cite{matusov-etal-2019-customizing} constrained the beam search to generate similar (source) length translations.
With regards to synchronizing translation with speech, \cite{federico_evaluating_2020} introduced a prosodic alignment model and later \cite{virkar2021pa} improved over that by utilizing speaking rate information and cross-lingual semantic matches to project source pauses to the target translation \yv{while} \cite{oktem2018heroes} \yv{leverages the attention weights in neural MT.} 
None of the previous works have looked at the problem of translation while maintaining speech synchronization as a whole except in a related work where \cite{karaknta2020subtitle} jointly learns to translate and project line breaks in the context of subtitling.

Despite the progress in adapting MT to use cases such as automatic dubbing, and subtitling, incorporating isochrony information in MT has not yet been explored. The main contributions of this work are:
\begin{itemize}
    \item Introduce the task of IAMT, investigate several approaches and report experimental results on a publicly available speech translation data set.
    
    \item Introduce a suite of automatic metrics to jointly evaluate phrase-pause alignment and verbosity of the translated phrases with respect to the source.
    
    \item Run subjective human evaluations on different MT system outputs and the final dubbed videos to measure the impact of the proposed approaches.
\end{itemize}

\begin{figure*}[t!]
    \centering
    \includegraphics[scale=0.99,trim={2.0cm 23.6cm 5cm 2.1cm},clip]{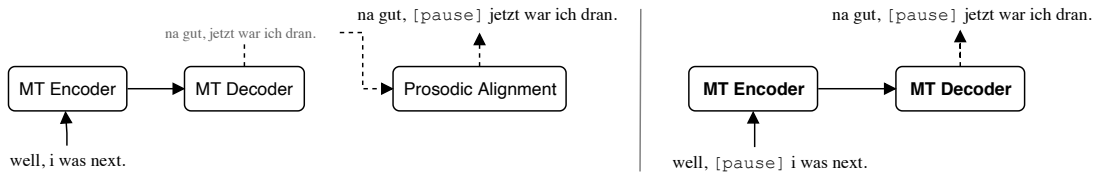}
    \caption{Two step approach of translation and pause projection using MT and prosodic alignment~\cite{federico2020ad} systems ({\it left}), and the proposed IAMT model generating translations along with {\tt [pause]} markers denoting the temporal speech pause ({\it right}).}
    \label{task}
\end{figure*}

% more examples
%  it's a question of [pas] freedom. [pas] against control 
%  es geht vielmehr um [pas] freiheit. [pas] oder kontrolle

\section{Isochrony-Aware Machine Translation}
\label{sec_task}
The task of IAMT involves translating sentences in source language containing pause markers correctly to the target language, which includes 1) \textit{projection of the pause markers} and 2) \textit{verbosity control of phrases} (see Fig. \ref{task}). To incorporate variable speaking styles, we refer to phrase as the text between two pauses, and not necessarily a group of words acting as a grammatical unit.%\footnote{We interchangeably use phrase and segment throughout the paper.} 
Below we will discuss our proposed approaches of implicitly and explicitly formulating IAMT.

\begin{table}[t]
\footnotesize
    \centering
    \tabcolsep=0.10cm
        \begin{tabular}{l c c c c }
        %\hline 
        \textbf{Method}     &  MT+PA    &MT+[pause]     &MT+DIS Emb      &MT+DIS Att  \\
        \hline
        SA       &100        &98.9           &30.1                       &30.1       \\
         %\hline
    \end{tabular}
    %}
    \caption{Preliminary results on En-De shows disentangled features (MT+DIS*) with embedding concatenation (Emb) or cross-attention (Att) under perform in terms SA metric.}
    \label{tab:pa_mt}
\end{table}

\subsection{Implicit Control of Pause Marker positions}
\subsubsection{Pause Marker (MT+{\tt [pause]})}
Injecting meta information or linguistic markers in neural MT is a well studied topic~\cite{sennrich-haddow-2016-linguistic,nadejde-etal-2017-predicting,hoang-etal-2016-improving}. A straightforward approach is to insert pause markers in the source and/or target text being generated. We simply add a {\tt [pause]} token to delineate pauses in the source and target sentences. %  and then let the MT model
The MT model learns the phrase boundaries implicitly leveraging the {\tt[pause]} positions along with tokens of the sequence. This way we incorporate pauses directly in the vocabulary of the model which allows it to learn semantics of the marker itself, and at the same time it implicitly learns to control the verbosity of the phrase by demarcating the phrase boundaries.

\subsubsection{Disentangling Feature (MT+DIS*)}
Factored MT has been used to inject external knowledge information in MT~\cite{garciafactored2018,wilkennovel2019} via a source factors~\cite{wilkennovel2019,dinu-etal-2020-joint} and/or as a target factors~\cite{tamchyna-etal-2017-modeling}. This factorization allows the model to disentangle the meta information from the actual input. In our case, we add a binary feature in both source and target side as separate factors indicating whether there is a pause marker after the current token. The model then use these disentangled features to output two predictions: the next token, and whether there is a pause marker after the next token. 

We experiment with two ways of modeling these two features: {\it i}) concatenating the embeddings of features and tokens in the input layer, and {\it ii}) having distinct encoders and decoders for tokens and binary features, connected by cross attention to model interactions between tokens and features.

We ran preliminary experiments with MT+{\tt [pause]} and the disentangling approaches (see Table \ref{tab:pa_mt}) against a strong baseline (MT+PA) where we first translate and then use a separate prosodic alignment model (with access to speech features) to project the pauses from source to target. These systems were evaluated on Over-/Under- Segmentation Accuracy (SA) i.e. accuracy of generating the same number of phrase segments in the target with respect to the source. SA is an initial indicator of the performance of the model on projection of pauses.
To circumvent the lack of labeled training data with [{\tt pause}] markers, we use a simulated training data for this process (refer Sec.~\ref{subsec:dataset} for more details). As it turns out, the segmentation accuracy goes down significantly for models with the disentangling feature, thus, we did not pursue this approach any further.

%\subsection{\task}
\subsection{Explicit Control of Pause Marker positions}
A drawback of the implicit approaches is the lack of control over the verbosity of each phrase between the pauses at inference time. While the projection of pause markers is important, we also need to control for the verbosity of phrases to achieve isochrony~\cite{sharma21b_interspeech}. 
As a result, we further consider the objective of controlling the verbosity of each phrase and as a by product, we can also maximize the SA.

In previous work \cite{takase_positional_2019,Lakew19}, verbosity control is implemented using length-dependent positional encoding. Motivated by this, we look at the problem of transferring pause markers as a modeling problem. The main difference from their work is that in ours ({\it MT+LC}) we have to control for verbosity at phrase level.

Similar to \cite{Lakew19}, we first compute the ratio of number of characters left to generate in the target sequence: $1 - \frac{\text{\# char\_generated}}{\text{\# total\_char}}$ 
where $total\_char$ is the length of the target phrase. %(pre-determined from the source length). 
These floating point ratios on intervals of $0.1$ from $0$ to $1$ are then quantized to an integer value between $0$ and $10$. The final embedding is the sum of token embeddings, sinusoidal positional embeddings~\cite{vaswani2017attention}, and the length dependent positional embeddings. 

During training, the model uses the total number of characters in the reference target phrase to compute the ratio of characters left to generate. 
At inference time, where we do not have references, the model uses the total number of characters in the source phrase to compute the ratio of characters left to generate. 
Different from~\cite{takase_positional_2019,Lakew19} which stopped generating at end-of-sequence token, we stop generating when the ratio of characters left to generate is zero. %This acts as a hard constraint, whereas previous models~\cite{Lakew19} use a soft constraint.

\section{Experiments}

\subsection{Dataset}
\label{subsec:dataset}
The HEROES data released by Oktem et. al. (2018)~\cite{oktem2018heroes} is the only publicly available data for the task of IAMT % isochrony-aware MT 
for English-Spanish, however, it contains 7,000 samples in total which renders it rather small to train MT model. 
MuST-Cinema data~\cite{karakanta2020cinemas} on the other hand has $\approx$200,000 samples for 7 language pairs but it is created specifically for speech to subtitles translation task, where segmentation is done on the basis of fixed length constraints (e.g. 42 characters), while for IAMT we require segmented input based on speech-pause information contained in the source audio.

MT model training with the proposed approaches require the source and target sentences to contain pause markers so that the model can learn to translate the content and project the pauses to the target side. However, since this information is unavailable from any of the open source data sets, we leverage the MuST-C speech-to-text translation data set~\cite{cattoni2020mustc}, to create training and gold standard test sets. 

\subsection{Datasets for IAMT}
\label{subsec:dataset_iamt}

\subsubsection{Training Data}
\label{subsubsec:training_data}
Training models for the IAMT task requires a much larger dataset (than the available ones) and performing forced alignment (for source pause markers) and post-editing (for target pause markers) at scale is very costly and time-consuming. To obtain source pause markers, a viable alternative is to use punctuation like comma or period characters. However, analyzing examples from \textit{MuST-C-495} (refer Sec~\ref{subsubsec:eval_data}), we noticed that the speakers do not necessarily pause in correspondence of these punctuation characters or other linguistic cues in the text. Rather, they pause at any point of time - for instance to catch their breath or get a glass of water during their talk.

To generate the training data, we insert the pause markers in the source text in the following way: \textit{First}, we computed a distribution of source phrase lengths from the source side of \textit{MuST-C-495} set. \textit{Second}, we randomly sampled a phrase length from this distribution and inserted pause markers after that desired phrase length. This way, we generated a phrase-pause structure in the source side of training data. 
To obtain target pause markers, we run a light weight PA module~\cite{virkar2021pa} (i.e. using only the cross-lingual semantic match features), to project pause markers from source to target text. 
In this way, for the \textit{train/dev} data sets we collected about 200,000 sentence pairs with pause markers synthetically generated in both the source and target languages for two directions, English-German/French. 

\subsubsection{Evaluation Data}
\label{subsubsec:eval_data}
For evaluation, we collected unique pairs of 495 sentences from the official MuST-C test set (which contain duplicate sentence pairs). Given the corresponding audio, \cite{virkar2021pa} annotated the pause information in the source sentence by force aligning the text with audio using Gentle aligner~\cite{ochshorn_gentle_2017}. For each phrase in the source sentence, the corresponding target phrase was post-edited with human annotators to create a parallel data with phrase-pause structure where the target phrases were similar in length to the source phrases.

\subsection{Models}
The baseline model (MT+PA) is a two step approach where we first translate the source text without pause marker using MT, and project the pause markers using the light weight PA module~\cite{virkar2021pa}. For IAMT task, we train models using the implicit (MT+{\tt [pause]}) and the explicit (MT+LC) approaches proposed in Sec.~\ref{sec_task}. 

Moreover, we compare the proposed models against the two step approach of Lakew et. al.~\cite{lakew2021verbosity}+PA approach, that trains MT with verbosity control, cascaded with the application of light-weight PA (as described in Sec \ref{subsubsec:training_data}). For all MT training we use the transformer base~\cite{vaswani2017attention} model configuration.

\subsection{Evaluation Metrics}
\label{subsec:eval}
Given that IAMT task is more complex than a standard translation task, we introduce additional metrics to measure three attributes: \textit{i}) translation quality at phrase level, \textit{ii}) segmentation accuracy, and \textit{iii}) length compliance across source and target phrases.

We measure overall translation quality (at corpus level) using detokenized BLEU~\cite{papineni2002bleu}, while at phrase level we evaluate translation quality with ChrF score~\cite{popovic-2015-chrf} (ChrF-Phrase) as precision for higher order $n$-grams might skew BLEU towards zero.  
To measure the accuracy of the projection of pauses over a data set, we compute the \% of sentences for which the number of pauses in the target is the same as in the source (SA, segmentation accuracy). 
For verbosity control, to measure length compliance at the phrase level, we consider the \% of sentences where length of every target phrase is within $\pm10\%$ range in character count of the corresponding (order wise) source phrase (PhraseLC). Implicitly, PhraseLC also takes into account that number of pauses on either side should be the same.

Finally, we compute a single score, that gives an overall picture of the three attributes in question: $Acceptability = ChrF{\text -}Phrase * PhraseLC$.

\begin{table}[t]
    \footnotesize
    \centering
    \tabcolsep=0.12cm
    \resizebox{0.5\textwidth}{!}{
    \begin{tabular}{l c c c c c | c  }
        %\hline 
        & \textbf{Method} & \textbf{BLEU}  & \textbf{ChrF-Phrase}  & \textbf{SA} & \textbf{PhraseLC} & \textbf{Acceptability} \\
        \hline
        \parbox[t]{2mm}{\multirow{3}{*}{\rotatebox[origin=c]{90}{En-De}}} & MT + PA &  27.5 & 58.5 & 100 & 16.1 & 9.6 \\
        & MT + [pause] & 27.8 & 59.5 & 99.8 & 19.7 & 11.7 \\
        & MT + \textsc{LC} & 26.5 & 50.4 & 100 & 39.1 & 19.7 \\
        & Lakew et. al.~\cite{lakew2021verbosity}+PA  & 28.8 & 51.2 & 100 & 43 & 22 \\
         \hline
        \parbox[t]{2mm}{\multirow{3}{*}{\rotatebox[origin=c]{90}{En-Fr}}} & MT + PA & 36.9 & 67.1 & 100 & 18.6 &  12.6  \\
        & MT + [pause] & 38.0 & 68.8 & 96.3 & 20.1 & 13.8   \\
        & MT + \textsc{LC} & 31.2 & 58.8 & 100 & 20.1 &  11.8  \\
        & Lakew et. al.~\cite{lakew2021verbosity}+PA & 38.4 & 60 & 100 & 43.1 & 25.8 \\
         \hline
    \end{tabular}
    }
    \caption{Results comparing the proposed IAMT approaches, MT+[pause] and \textsc{MT+LC} against the cascaded baseline MT+PA and the current best MT with verbosity control mechanism of Lakew et. al.~\cite{lakew2021verbosity}+PA, on \textit{MuST-C-495} test set \cite{virkar2021pa}.
    }
    \label{tab:pa_mt_vebosity}
\end{table}

\section{Results: Automatic Evaluation}
Table~\ref{tab:pa_mt_vebosity} collects results for all systems evaluated on \textit{MuST-C-495} post-edited test set for both En-De and En-Fr language pairs. 
Looking at the \textit{Acceptability} scores, the approach of Lakew et. al.~\cite{lakew2021verbosity}+PA achieves the best results but this is expected as it first applies MT and a re-ranking module to control for verbosity, and then leverages a PA module (trained with speech features) specifically for phrase segmentation. Our aim is not to improve over this cascaded system rather get as close as possible without deploying multiple modules into production.

The most interesting finding is that while MT+LC outperforms the MT+PA on length compliance of phrases it does so at trade-off with translation quality (c.f. ChrF-Phrase). This is expected because MT+LC optimizes on phrase level verbosity control. MT+{\tt [pause]} on the other hand, consistently fares better against a strong baseline of MT+PA in terms of ChrF-Phrase, PhraseLC and Acceptability score. This means an implicit way of integrating pause markers into MT provides a better trade-off on all three attributes.

\begin{table}[t]
\footnotesize
    \centering
    \footnotesize
    \tabcolsep=0.17cm
    % \resizebox{0.5\textwidth}{!}{
    \begin{tabular}{l c c c c}
        %\hline 
        & \textbf{Method}  & Acceptable  & Fixable & Wrong  \\
        \hline
        \parbox[t]{2mm}{\multirow{3}{*}{\rotatebox[origin=c]{90}{En-De}}} 
        & MT + [pause]  & 26.2 & 35.3 & 38.5   \\
        & MT + \textsc{LC}  & 12.1 & 29.0 & 58.9  \\
        & Lakew et. al.~\cite{lakew2021verbosity}  & 26.8 & 36.7 & 36.5 \\
         \hline
        \parbox[t]{2mm}{\multirow{3}{*}{\rotatebox[origin=c]{90}{En-Fr}}} 
        & MT + [pause]  & 26.5 & 35.8 & 37.69 \\
        & MT + \textsc{LC}  & 8.2 & 28.2 & 63.6 \\
        & Lakew et. al.~\cite{lakew2021verbosity}   & 31.9 & 41.2 & 26.9 \\
         \hline
    \end{tabular}
    % }
    \caption{Human evaluation of MT system outputs (without pauses) on a $200$ randomly selected unique samples from the post-edited benchmark.}
    \label{tab:mt_human_eval}
\end{table}
\begin{table*}[t]
\footnotesize
    \centering
    \begin{tabular}{c|c|c|c|c|c|c|c}
        \hline 
        \textbf{(I)} &  & A & B & C & D & B$^{'}$ & D$^{'}$ \\
        %\textbf{(I)} &  & MT+PA & MT+[pause] & MT+LC & VC+PA & MT+[Pause]$^{'}$ & VC+PA$^{'}$ \\
        \hline
        En-De & Smoothness & 51.9 & 56.3 & 65.6 & 48.4 & 56.6 & 55.9\\
        En-Fr & Smoothness & 44.8 & 53.1 & 60.0 & 40.0 & 55.2 & 53.3\\
        \hline
    \end{tabular}
    \begin{tabular}{l|l|ll|ll|ll||ll || ll}
    \hline
      \textbf{(II)}&    & \multicolumn{2}{l|}{A\hspace{0.5cm}   vs.   \hspace{0.3cm} B} &    \multicolumn{2}{l|}{A\hspace{0.5cm}   vs.   \hspace{0.3cm} C} & \multicolumn{2}{l||}{B\hspace{0.5cm}   vs.   \hspace{0.3cm} C} & \multicolumn{2}{l||}{D\hspace{0.5cm}   vs.   \hspace{0.3cm} B} & \multicolumn{2}{l}{D$^{'}$\hspace{0.5cm}   vs.   \hspace{0.3cm} B$^{'}$} \\
      \hline
      %Wins  \hspace{0cm} & 28.37\% & 46.33\% ** & 29.9\% & 42.24\% ** \\
     En-De & Wins  \hspace{0cm} & 32.0 & \hspace{0.2cm} 41.0$^*$ & 48.4 & \hspace{0.2cm}30.8$^*$ & 51.7 & \hspace{0.2cm}30.1$^*$ & 34.8 & \hspace{0.2cm}40.9$^+$ & 37.4 & \hspace{0.3cm}37.5\\
     En-Fr & Wins  \hspace{0cm} & 36.9 & \hspace{0.3cm}38.2 & 61.4 & \hspace{0.2cm}25.8$^*$ & 60.9 & \hspace{0.2cm}22$^*$ & 29.9 & \hspace{0.2cm}43.4$^*$ & 45.0 &\hspace{0.2cm} 35.1$^*$\\
    \hline
    \end{tabular}
    \caption{\textbf{(I)} Automatic smoothness metric \cite{virkar2021pa} and \textbf{(II)} Subjective user preferences (\% of Wins) for automatic dubbing in a head to head comparison of: (A) MT+PA, (B) MT+[pause], (C) MT+\textsc{LC}, and (D) Lakew et al. \cite{lakew2021verbosity}+PA. Models B$^{'}$, D$^{'}$ are versions of models B, D that also apply the relaxation mechanism in \cite{virkar2021pa}.  Significance testing is done for the Wins %metric 
    with levels $p < 0.05$ ($^+$) and $p < 0.01$ ($^*$).}
    \label{tab:ADresults}
\end{table*}

\section{Results: Human Evaluation}
\subsection{Machine Translation Evaluation}
\label{subsec:mt_eval}
Human evaluation follows a simple yet an effective strategy to grade both quality and fluency of the MT outputs. 
Following~\cite{lakew2021isometricmt}, we ask subjects to rate 200 randomly selected translation subset of the test as acceptable, fixable, or wrong with respect to the reference.\footnote{\textit{Acceptable}: meaning is similar, fluency is good, \textit{Fixable}: meaning is similar, fluency is poor, and \textit{Wrong}: meaning is different.}

Table~\ref{tab:mt_human_eval} show results comparing two of our proposed approaches against the state-of-the-art Lakew et al. \cite{lakew2021verbosity} for MT verbosity control. %, that leverages N-best list re-scoring. 
For En-De, MT+{\tt [pause]} shows comparable performance with acceptable translations at $26.2\%$ with respect to~\cite{lakew2021verbosity} at $26.8\%$. For En-Fr, MT+{\tt [pause]} drops by $5.4\%$ from the best performing~\cite{lakew2021verbosity}. For the MT+LC model, we observed a large drop in the acceptable translations, which we regard as the outcome of an aggressive verbosity control that pushes the model to drop certain tokens. 
Overall, from the MT human evaluation we confirm that implicitly modeling the pause information with MT is a promising direction for Isochrony aware MT.

\subsection{Dubbing Evaluation}
We present results of human evaluation on a random subset of 50 single-sentence test videos. For each source sentence and corresponding video clip, we create dubbed videos using the dubbing architecture proposed by~\cite{federico_evaluating_2020} with the following systems: (A) MT+PA, (B) MT+[{\tt pause}], (C) MT+LC and (D) Lakew et al. \cite{lakew2021verbosity} + PA. For the two systems in which no separate PA module is applied (B and C), time stamps of the source pauses are directly projected to the corresponding target pauses. For all systems, TTS audio is generated according to the duration of each target segment in order to fit the speech timing of the video. To reduce the cognitive load, we conduct separate  evaluations comparing only two systems at a time. For all evaluations, we show as a reference a dubbed video generated from manually post-edited and segmented translations. Human subjects first watch the reference dubbed video and then rate viewing experience of videos dubbed with two systems on a scale of 0 to 10 with 10 being the highest quality and 0 being the worst quality. 

We run evaluations on En-De and En-Fr directions with 40 human annotators, who are native speakers in the target language, with each of them grading 25 of the 50 videos, resulting in a total of 1,000 data points for each comparison. We report Wins, i.e., the \% of times one system is preferred over the other. 

Part \textbf{(I)} of Table~\ref{tab:ADresults} shows results for automatic evaluation with the Smoothness metric \cite{virkar2021pa} that computes the stability of TTS speaking rate across contiguous target phrases. Part \textbf{(II)} shows the results for subjective human evaluation with the Wins metric. For both automatic and human metrics system B outperforms system A on both languages with relative improvements for Smoothness (De: +8.5\%, Fr: +18.5\%) and Wins (De: +28.1\%, Fr: +3.5\%) with statistically significant ($p < 0.01$) difference in Wins for De. B significantly outperforms C on both languages in terms of Wins (De: +57.1\%, Fr: +137.9\%). Though C has the better Smoothness compared to B, as shown in Sec.~\ref{subsec:mt_eval}, C trades off on translation quality for improved Smoothness and hence results in automatically dubbed videos of lower quality.

\subsubsection{Relaxation Mechanism}
Comparing system D (with light weight PA, without speech features) with B, the latter is better on both Smoothness (De: +16.3\%, Fr: +32.8\%) and significant Wins (De: +17.5\%, Fr: +45.2\%) for both languages. However, this result does not take into account the relaxation mechanism~\cite{virkar2021pa} that can improve speaking rate smoothness.

Therefore, we applied the relaxation on the above two system and denote with B$^{'}$ and D$^{'}$ dubbing obtained with outputs from B and D after applying the relaxation. Note that, for D$^{'}$ we apply a full fledged PA module (with speech features), while B$^{'}$ is devoid of any such PA module. From Part \textbf{(I)}, we observe that Smoothness of D$^{'}$ and B$^{'}$ is improved as expected compared to D and B.  Additionally Smoothness of D$^{'}$ and B$^{'}$ are now comparable. Also, D$^{'}$ beats B$^{'}$ for Wins on Fr (+28.2\%, $p<0.01$) and obtains comparable Wins for De. The reason is that after adding relaxation, while both systems reach comparable smoothness, D actually provides more acceptable translations than B (cf. Table~\ref{tab:mt_human_eval}). In fact, in order to generate high quality dubs, both translation quality and speaking rates are necessary components, and trading-off between these is the main challenge for IAMT.

\section{Conclusion}
In this work, we introduced an isochrony-aware MT task, where one has to transfer pause information from  source to target along with translating the content. We proposed metrics to evaluate on multiple attributes; the correct number of pause markers, their positions, and verbosity at the level of phrase segments. We compared our proposed approaches (to model pause positions and translation) against strong baseline systems that decouples MT and prosodic alignment steps. We conducted automatic and human evaluations both on translation quality and on automatic dubbing, which relies on prosodic and temporal information projected from the source. As it turns out, the best approach to model both pause information and translation is to simply inject the pause markers in the text and let the model implicitly learn the two tasks.

\bibliographystyle{IEEEbib}
%\bibliography{strings,refs}
\bibliography{paper}

\end{document}